\documentclass[a4paper]{article}

\usepackage{INTERSPEECH2020}
\usepackage{amsmath,graphicx, tabularx, xcolor}
\usepackage{multirow}
\usepackage{multicol}
\usepackage{hyperref}
\usepackage{subcaption}
\usepackage{footnote}

\newcommand{\prgname}{ClovaCall}
\newcommand{\prgnameb}{ClovaCall-Base}
\newcommand{\prgnamef}{ClovaCall-Full}
\newcommand{\aicc}{AICC}

\title{ClovaCall: Korean Goal-Oriented Dialog Speech Corpus \\for Automatic Speech Recognition of Contact Centers}
\name{Jung-Woo Ha$^{1}$\sthanks{The first two authors equally contributed to this work.}, Kihyun Nam$^{1,2*}$, Jingu Kang$^1$, Sang-Woo Lee$^1$, Sohee Yang$^1$, \\ Hyunhoon Jung$^1$, Hyeji Kim$^1$, Eunmi Kim$^{1}$, Soojin Kim$^1$, Hyun Ah Kim$^1$, \\ Kyoungtae Doh$^1$, Chan Kyu Lee$^1$, Nako Sung$^{1}$, Sunghun Kim$^{1}$}
\address{
  $^1$Clova AI, NAVER Corp., 
  $^2$Hankuk University of Foreign Studies}
\email{jungwoo.ha@navercorp.com}

\begin{document}

\maketitle
\begin{abstract}
Automatic speech recognition (ASR) via call is essential for various applications, including AI for contact center (\aicc{}) services. Despite the advancement of ASR, however, most publicly available call-based speech corpora such as Switchboard are old-fashioned. Also, most existing call corpora are in English and mainly focus on open domain dialog or general scenarios such as audiobooks. Here we introduce a new large-scale Korean call-based speech corpus under a goal-oriented dialog scenario from more than 11,000 people, i.e., \prgname{} corpus. \prgname{} includes approximately 60,000 pairs of a short sentence and its corresponding spoken utterance in a restaurant reservation domain. We validate the effectiveness of our dataset with intensive experiments using two standard ASR models. Furthermore, we release our~\prgname{} dataset and baseline source codes to be available here~\footnote{\textcolor{blue}{\small{\url{https://github.com/ClovaAI/ClovaCall}}}}.
\end{abstract}
\noindent\textbf{Index Terms}: \prgname, Korean call speech corpus, automatic speech recognition, goal-oriented dialog utterance

\section{Introduction}

\label{sec:intro}
Call-based customer services are still prevalent in most online and offline business. In particular, call centers have played a crucial role in most business domains for a few decades and recently extended to contact centers which provide additional functions such as email, VoIP, and text chatting\footnote{\url{https://aircall.io/blog/call-center/contact-center-vs-call-center/}}. However, the increasing costs and the harsh working environments of contact centers have brought the necessity to apply artificial intelligence (AI) to contact center operation~\cite{delgado2020emerging}. AI for contact center (\aicc) is an AI agent that communicates with human customers via call, which rapidly increases in B2B markets~\cite{delgado2020emerging}. Since \aicc{} is based on a telephone environment, automatic speech recognition (ASR) via call is essential for successful \aicc{} operation. 

ASR has been one of the tasks remarkably improved by deep learning since early years of 2010s~\cite{yao2012adaptation,mohamed2011acoustic}. 
It is well known that the improvement of ASR results from large-scale speech corpora including Wall Street Journal~\cite{paul1992design}, TIMIT~\cite{garofolo1993timit},  Switchboard~\cite{godfrey1997switchboard}, CallHome~\cite{canavan1997callhome}, and Librispeech~\cite{panayotov2015librispeech} datasets.   
However, most publicly available call speech corpora are very old-fashioned such as Swichboard, Wall Street Journal, and CallHome because they were released more than 20 years ago. Also, the language of most call corpora is mainly English, and thus call corpora of low resource languages are very scarce. The other issue is is that the utterances of most corpora are based on open domains such as day-life conversation and audiobook contents. Even if small numbers of Korean speech corpora are publicly available such as AIHub~\footnote{\url{http://www.aihub.or.kr/aidata/105}} and Zeroth project\footnote{\url{https://github.com/goodatlas/zeroth}}, they contain general open domain dialog utterances. Therefore, ASR models trained from these speech data generally show poor recognition performance when applied to domain-specific tasks due to the differences in their data distribution and vocabularies. In particular, \aicc{} requires an accurate ASR model to ensure the precise intent classification or slot extraction~\cite{chen2019bert} from user natural language utterances.

\begin{figure*}[!t]
\begin{center}
\includegraphics[width=0.9\textwidth]{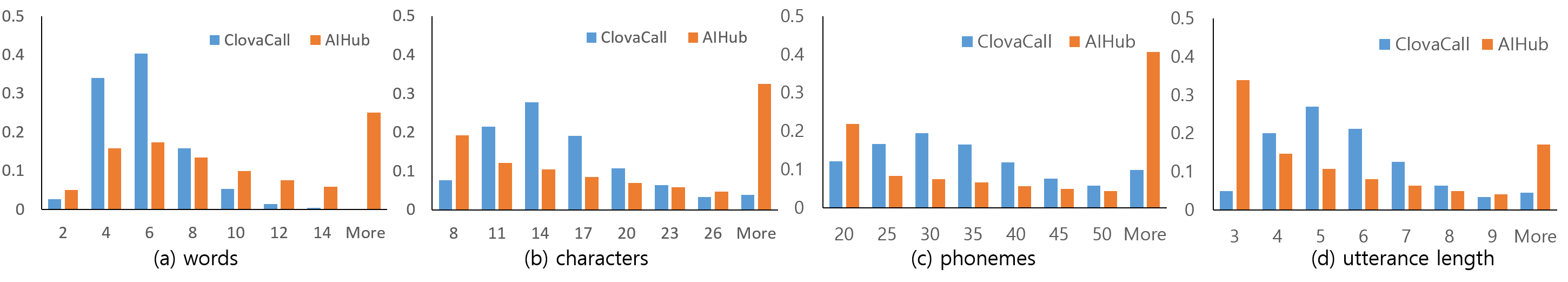}
\end{center}
\vspace{-0.5 cm}
   \caption{Distributions of four attributes on the \textit{raw} set of \prgname{} dataset (goal-oriented) with blue bars compared to those of AIHub dataset (open domain) with orange bars. We can find the different patterns between two datasets for all attributes.}
\label{fig:graph_stat}
\end{figure*}

Here we release a new large-scale Korean call speech corpus containing goal-oriented dialog utterances under a restaurant reservation scenario, i.e., \prgname{} speech corpus. The proposed \prgname{} includes 61,000 pairs of short sentences and their utterances recorded via call by more than 11,000 people. In specific, the number of unique sentences is 8,990, and all of them are natural language questions and answers which frequently appear when making reservations. The utterances that each subject read given sentences aloud are recorded over a phone. Because most sentences are designated for reservation and short with at most 10 seconds, our dataset does not suffer from end point detection and alignment problems, dissimilar to Librispeech which is extracted from audiobooks. \prgname{} can be useful for diverse \aicc-based reservation services because most words and expressions prevalent in reservations are commonly used regardless of its application domains, including time, people, date, and location. 

We demonstrate the effectiveness of the proposed \prgname{} with extensive experiments. We employ two standard ASR models such as Deep Speech 2 (DS2)~\cite{amodei2016deep} and Listen, Attend and Speech (LAS)~\cite{chan2016listen} under three training schemes including pretraining-finetuning, from-scratch training, and scratch training with data augmentation. Besides, we use two additional datasets for effective verification. One is an in-house Korean call-based goal-oriented dialog speech corpus on questions and answers for daily company life (QA Dataset) for verifying the necessity of task-specific data. The other is a large-scale Korean open domain dialog speech corpus from AIHub, an online Korean data hub site, for pretraining ASR models. 
Experimental results show the ASR models trained from large-scale open domain data only provide very poor recognition performances. Thus, the task-specific speech datasets are essential for speech recognition of goal-oriented dialogs like \aicc. Interestingly, pretraining with open domain data remarkably improves the ASR accuracy compared to scratch training with task-specific data only even though their sampling rates and frequently used words are different from each other. 


\section{Related Work}
\label{sec:rw}
Large-scale speech corpora publicly available allows ASR models to be applied to many valuable real-world applications. Early public speech corpora were released in 1990s, including Wall Street Journal~\cite{paul1992design}, TIMIT~\cite{garofolo1993timit},  Switchboard~\cite{godfrey1997switchboard}, and CallHome~\cite{canavan1997callhome}. These datasets are still prevalent as benchmark datasets for evaluating ASR models~\cite{amodei2016deep,saon2017english,ravanelli2018light,baevski2019vq}. More recently, Librispeech~\cite{panayotov2015librispeech} is the most popular benchmark speech corpus on which the latest state-of-the-art ASR models are evaluated~\cite{han2019state,luscher2019rwth,park2019specaugment,baevski2020effectiveness}. Despite their usefulness, existing speech corpora mainly deal with general open domain dialogs. Even if the large-scale corpora are helpful for pretraining ASR models, the models not finetuned with task-specific data are likely to provide poor recognition accuracy when applied to recognize user utterances in goal-oriented scenarios such as call centers and reservations services (See Sec.~\ref{sec:results}). This poor performance results from the distribution difference between open domain and task-specific goal-oriented dialogs. However, compared to open domain dialog speech corpora, goal-oriented speech corpora are rarely released publicly. 

\section{Clova Call Speech Corpus}
\label{sec:datset}

\subsection{AI for Contact Center}
\prgname{} dataset construction is one of main subtasks in our \emph{AI for Contact Center (AICC)} project~\footnote{\textcolor{blue}{\small{\url{https://clova.ai/aicontactcenter}}}}. The goal of \aicc{} is to develop an AI agent which can help human contact center employees to communicate with customers via phone. In perspective of technology, the main functionality of \aicc{} contains ASR, natural language understanding such as intent classification and slot filling, goal-oriented style dialog management, response generation, and voice synthesis. Here, we focus on its ASR component and construct a large-scale speech corpora concentrating on a restaurant reservation scenario. 

\subsection{Data Construction from Humans}
\prgname{} contains 60,746 utterance and short sentence pairs on the restaurant reservation scenario via call. The process of data construction was carried out in the following order: 1) making a sentence pool, 2) call-based recording utterances with the sentences, and 3) refining the recorded speech data. 

\noindent\emph{\textbf{Sentence pool. }} We utilized Crowdworks\footnote{\url{https://www.crowdworks.kr/}}, a Korean crowd sourcing platform, to make a pool of candidate sentences.
First, we defined 10 categories, 86 intents, and 7 multi-turn situations for restaurant call scenarios.
10 categories, which are high-level topics, include \texttt{reservation}, \texttt{delivery}, and 8 FAQ categories like \texttt{working time}, \texttt{menu}, and \texttt{discount}.
86 intents belong to one of 10 categories and contain \texttt{whether the restaurant is opened now}, \texttt{closing time}, \texttt{recommended menu}, etc.
We also defined 7 multi-turn situations, which could be appear in a call to restaurants, inclu \texttt{reservation change} and \texttt{delivery call}, etc.
The crowd-workers were asked to imagine and generate multiple interrogative or answer sentences for given intents and situations. After quality assurance process was performed by human experts, 8,990 sentences, which are mainly answer sentences, were selected to comprise the candidate pool by eliminating duplicated sentences.


\begin{table}[t!]
\small\centering
\begin{tabular}{l|rrr|r}
\hline
\hline
Values & Words & Chars & Phonemes & Utter. time \\
& & & & + silenence \\ \hline
Voc size & 4,704 & 613 & 53 & - 
\\ 
\hline
Mean & 4.39 & 13.79 & 32.39 & 2.94s
\\ & & & & +2.57 \\ \hline
Stdev & 1.99 & 5.50 & 12.99 & 1.77s
\\ & & & & +0.79 \\ \hline
Max / Min & 17 / 1 & 48 / 3 & 116 / 5 & 30s / 0.3s 
\\ & & & & + 0 / 0.7 \\ \hline\hline
\end{tabular}
\caption{Statistics of four attributes of \prgnameb{} dataset. silence means including silence regions in utterances}
\label{table:data_stats}
\vskip -0.25in
\end{table}

\noindent\emph{\textbf{Call-based recording utterances. }} 
Utterance recording was performed based on crowd sourcing, operated by ourselves. 10 unique sentences are given to each crowd-worker. The crowd-worker reads each of the sentences aloud once or twice via call to make at most 20 utterances, which were transmitted into our server. From 11,000 people, we gathered more than 120k pairs of short sentences and utterances. Compared to Librispeech, there do not exist end point detection and alignment problems in our data because the utterances are short enough considering call-based reservation scenarios. Besides sentences, each utterance also has its anonymous speaker index as one of the labels. This allows our data to be useful for speaker identification task. 

\noindent\emph{\textbf{Refining data. }} 
Data gathered via crowdsourcing are likely to contain many noises, and thus it is essential to refine the gathered data. First, we carried out qualitative evaluation on the gathered data, which was performed by human experts engaged in CrowdWorks so that we could select a total of 82,306 utterance-sentence pairs. This is the \textit{raw} version of \prgnamef{}. Next, we removed the starting or the ending silence regions below a specific energy level in the raw waveform of utterances. We used \texttt{librosa}~\cite{mcfee2015librosa} with 25db as the threshold for silence elimination. The silence-free data is called \textit{clean} version. Finally, we selected top-30 intents with the most utterance-sentence pairs to be a dataset containing 60,746 pairs. We call the dataset \prgnameb{}. We release the \textit{clean} version of \prgnameb{} via \textcolor{blue}{\small{\url{https://github.com/clovaai/ClovaCall}}}. From now, \prgname{} denotes to the \textit{clean} version of \prgnameb{} for convenience.

\begin{figure}[!t]
\begin{center}
    \includegraphics[width=0.9\columnwidth]{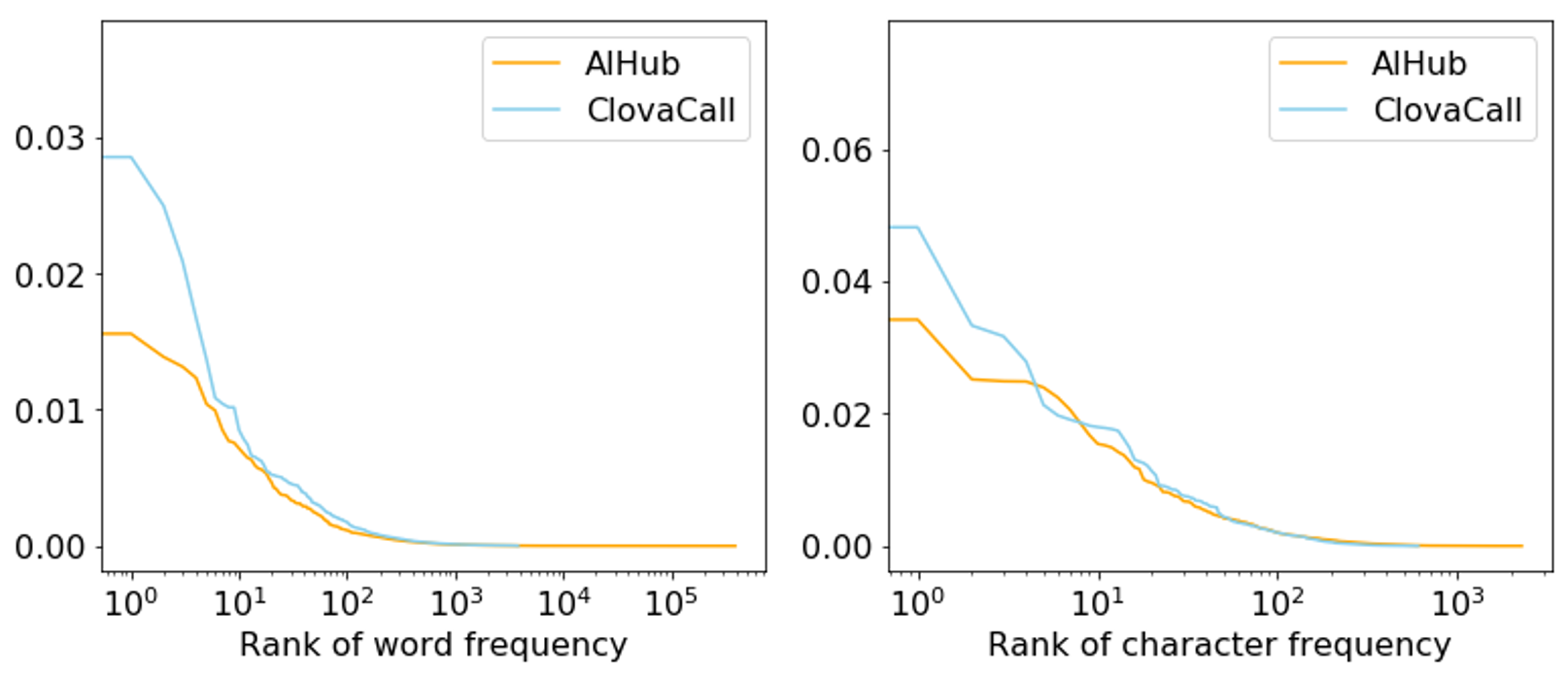}
\end{center}
\vskip -0.25in
   \caption{Comparison between ClovaCall and AIHub datasets in terms of frequency rank-ratio of each word and character. Left and right graphs are for word and character.}
   \label{fig:unit_dist}
\end{figure}

\begin{table}[t!]
\small\centering
\begin{tabular}{c|r|r|r|r}
\hline
\hline
ClovaCall & Ratio & ~Size~ & ~ No ~ & Rank in AIHub \\
\hline
\multirow{4}*{Words} & 0.1 & 381 & 56 & 28,979 \\
& 0.25 & 952 & 230 & 37,664 \\
& 0.5 & 1,904 & 662 & 45,459 \\
& 1.0 & 3,808 & 1,714 & 52,065\\
\hline
\multirow{4}*{Chars} & 0.1 & 61 & 0 & 93.7 \\
& 0.25 & 151 & 0 & 162.8 \\
& 0.5 & 302 & 0 & 227.3 \\
& 1.0 & 603 & 3 & 388.3 \\
\hline
\hline
\end{tabular}
\caption{Usage patterns of top frequent words and characters of \prgname{} in AIHub dataset. Ratio and size denote the ratio and the number of the top ranked words and characters. No means the number of words and characters not used in AIHub. Rank in AIHub is the average rank in AIHub. The unique word and character sizes are different from Table~\ref{table:data_stats} because we discarded numerical characters. The sizes of unique words and characters of AIHub are 390,599 and 2,268.}
\label{table:overlap}
\vskip -0.25in
\end{table}

\subsection{Analysis on ClovaCall Dataset}
To show the difference from open domain speech corpus, we compared our dataset with AIHub dataset, the largest Korean open domain dialog corpus. Fig.~\ref{fig:graph_stat} illustrates the frequency histograms of words, characters, phonemes, and utterance length. Overall, most sentences in \prgname{} include more than 4 and less than 8 words , and more than 11 and less than 20 characters. Thus, the length of most utterances is more than 4 and less than 10 seconds. These distributions reveal the characteristics of restaurant reservation scenario. Compared to those of AIHub, the frequencies of each attribute are more concentrated on a specific region. We conjecture this pattern results from that most utterances in \prgname{} are likely to contain information for reservation while open domain dialog covers much more diverse topics and situations including both very short response utterance such as \textit{``Yes''} and \textit{``Sure''}, and long utterances. Table~\ref{table:data_stats} depicts the number of unique elements, mean, standard deviation, maximum and minimum values for word, character, phoneme, and utterance length. Interestingly, the mean values of utterance length and silence time are very similar, which is caused by call-based recording setup. 

Fig.~\ref{fig:unit_dist} depicts the difference of frequency ratio of each word and character frequency rank. As shown in Fig.~\ref{fig:unit_dist}, we can find that higher ranked words and characters of \prgname{} are used much more frequently than those of AIHub. Table~\ref{table:overlap} shows the differences of usage pattern between ClovaCall and AIHub. In terms of characters, \prgname{} shows the similar pattern to AIHub considering the number of not used characters and their average rank in AIHub. However, the word usage of \prgname{} is significantly different from AIHub. Frequently used words of \prgname{} do not appear in AIHub or are rarely used with very low rank. These difference between two datasets enhances the necessity of goal-oriented dialog corpora. We show the difference causes poor accuracy of ASR models trained from AIHub data on ClovaCall dataset. 

\section{Speech Recognition Results}
\label{sec:asr_result}
\subsection{Experimental Setup}

\noindent\emph{\textbf{Datasets. }}
We use two additional datasets besides \prgname{} to effectively verify the efficacy of our dataset. One is our in-house speech corpus on internal questions and answers about company lives collected via phone calls, called QA Call dataset. The other is a large-scale Korean open domain speech dialog corpus from NIA AIHub, an open data hub site of Korea Government. The AIHub speech is used for pretraining the ASR models. Also, we verify the results on \prgnamef{} in addition to \prgname{}. While QA Call and ClovaCall are sampled with 8kHz, AiIHub contains the speech voices recorded with 16kHz sampling rate. As shown in Table~\ref{table:data_desc} for experiments, we separate 59,662 and 1,084 sentence-utterance pairs from \prgnameb{} as training and test sets. The training set of \prgnamef{} contains approximately 22,000 more pairs whose intent is excluded from \prgnameb{}. There is neither duplicated speaker nor sentence between two separated sets. For QA Call, we extract the same size of sentence-utterance pairs as \prgname{} as the training set. More data were used as test set of QA Call for robust evaluation. For fair comparison, the augmented amount is similar to the pretraining data of AIHub, which is explained in the next section. In addition, the finetuning data size of AIHub is equal to the training data size of two goal-oriented datasets.

\begin{table}[t!]
\small\centering
\begin{tabular}{c|c|r|r}
\hline
\hline
Dataset~~~~&~~~~~Type~~~~~&~~~~Number~~~~&~~~~Hour~~~~ \\
\hline
\multirow{4}*{\prgname{}}& \underline{Training-Base} & \underline{59,662} & \underline{50}\\
& Training-Full & 81,222 & 67 \\
& Noise Aug & 406,110 & 337 \\
& \underline{Test} & \underline{1,084} & \underline{1} \\
\hline
\multirow{3}*{QA Call}& Training & 80,984 & 83 \\
& Noise Aug & 404,920 & 415 \\
& Test & 10,000 & 12.4 \\
\hline
\multirow{2}*{AIHub}& Pretraining & 381,603 & 510 \\
& Finetuning & 80,105 & 100 \\
\hline\hline
\end{tabular}
\caption{Description of three used datasets for experiments. Underlined data were released as \prgname{} dataset.}
\label{table:data_desc}
\vskip -0.25in
\end{table}

\noindent\emph{\textbf{Training schemes. }} For verifying the effectiveness of \prgname{} and the necessity of task-specific speech corpus, we employ three training scenarios: 1) pretraining and finetuing, 2) training from scratch, and 3) training from scratch with data augmentation. AIHub dataset is used for pretraining. Also, almost the same amount of AIHub data to the training portion of \prgname{} and QA Call was used for finetuning to investigate whether call-based goal-oriented utterance data are essential or not for task-specific services. We verify the effectiveness of pretraining with open domain speech corpora by comparing the results to those trained from data enhanced by two data augmentation methods such as noise augmentation and specaugment~\cite{park2019specaugment}. We augmented data with noises using our in-house room simulator by adding different types of noise and reverberations~\cite{ko2017study} that we obtained from daily environmental recordings. We did not perform experiments with a language model to enhance the ASR accuracy because we mainly focus on the efficacy our dataset in goal-oriented scenarios.

\noindent\emph{\textbf{Data preprocessing. }} First, we upsample the 8Khz waveform datasets to 16Khz so that all datasets have the same frequency resolution. The reason of using upsampling instead of downsampling is that we assume an environment where conventional ASR models for 16KHz sampling rate are used for recognizing call-based speech signals. All models use log-spectrograms as input data, which are calculated with 20ms window size and 10ms stride size using \texttt{librosa}. In addition, all spectrograms were normalized by instance-wise standardazation.

\noindent\emph{\textbf{ASR Models. }} We use two standard ASR models such as DS2~\cite{amodei2016deep} and LAS~\cite{chan2016listen} for verifying the effectiveness of our proposed \prgname{}. DS2 consists of a CNN and an RNN. In our setting, the CNN module has two 2D-Convolutional layers with 32 channels, which reduce both the frequency and the time resolution of the input spectrogram with stride 4 and 2 for each layer. The RNN module consists of five bidirectional LSTM layers. All these layers have 800 hidden units per direction, in total, 1600 units per layer. Next, one fully connected layer outputs the softmax distribution over characters. Finally, DS2 is trained with CTC loss~\cite{graves2006connectionist}. More details of DS2 are described in \cite{amodei2016deep}. LAS is a sequence-to-sequence model consisting of an encoder, decoder, and attention. The encoder includes a CNN module and an RNN module sequentially. The CNN module is identical to that of DS2. The RNN module of LAS encoder consists of three stacked bidirectional LSTMs with 512 units per direction. The decoder has two unidirectional LSTMs with 512 units and one fully connected layer to predict the character probability distribution. The attention learns the alignment between the encoder outputs and the decoder hidden states. Location-aware attention~\cite{chorowski2015attention} is employed for the attention context of the previous state. All the experiments are performed based on NAVER Smart Machine Learning (NSML) platform~\cite{kim2018nsml,sung2017nsml}.

\noindent\emph{\textbf{Metrics. }}
We use character error rate (CER) as a metric:
\[D = Distance_{LEV}(X,Y),~ CER(\%) = \frac{D}{L} \times 100\]
\noindent where $X$, $Y$ are a predicted and a ground truth scripts. The distance $D$ is the Levenshtein distance between $X$, $Y$~\cite{yujian2007normalized} and the length $L$ is a length of ground truth script $Y$.

\begin{table}[t!]
\small\centering
\begin{tabular}{l|rr|rr}
\hline
\hline
\multirow{2}*{Models }&\multicolumn{2}{c|}{DS2~\cite{amodei2016deep}}&\multicolumn{2}{c}{~~~LAS~\cite{chan2016listen}~~~} \\
(Parameters)&\multicolumn{2}{c|}{(56M)}&\multicolumn{2}{c}{(31M)} \\
\hline
Data &~~~$QA$~~~ &~~~$CC$~~~&~~~$QA$~~~ &~~~$CC$~~~ \\
\hline
\multicolumn{5}{c}{Pretraining and finetuning} \\
\hline
$A^{pt} \rightarrow A^{ft}$~~~& 54.6 & 59.5 & 62.3 & 69.2 \\
$A^{pt} \rightarrow QA$~~~& 12.2 & 25.6 & \textbf{10.9} & 26.7 \\
$A^{pt} \rightarrow CC$-Base~~~& 35.9 & 9.54 & 38.7 & 8.0 \\
$A^{pt} \rightarrow CC$-Full~~~& 32.9 & 8.31 & 35.3 & \textbf{7.0} \\
\hline
\multicolumn{5}{c}{From-scratch training} \\
\hline
$QA$ & 15.2 & 34.7 & 15.3 & 40.0 \\
$CC$-Base & 75.2 & 16.7 & 87.7 & 22.1 \\
$CC$-Full & 62.3 & 11.4 & 76.7 & 15.1 \\
\hline
\multicolumn{5}{c}{From-scratch training with data augmentation} \\
\hline
$QA$ /w NA & 16.5 & 38.5 & 14.8 & 38.3 \\
$CC$-Full /w NA & 64.4 & 10.7 & 81.4 & 18.9 \\
$QA$ /w SA & 16.5 & 39.7 & 17.1 & 43.5 \\
$CC$-Full /w SA & 63.4 & 10.1 & 88.3 & 31.1 \\
\hline
\hline

\end{tabular}
\caption{CER of each ASR model on two datasets under three training schemes. $A$, $QA$, $CC$ denote AIHub, QA Call, \prgname{} datasets, respectively. $pt$ and $ft$ mean the data for pretraining and finetunig. NA and SA are noise augmentation~\cite{ko2017study} and specaugment~\cite{park2019specaugment} on \prgnamef{}. }
\label{table:user_study}
\vskip -0.25in
\end{table}

\subsection{Comparison Results on Datasets}
\label{sec:results}
Our experiments focus on verifying the effectiveness of task-specific speech corpora for a certain \aicc{} services. Table~\ref{table:user_study} depicts the results of two popular ASR models under the three training scenarios described in Sec 4.1. In the pretraining and finetuning scheme, despite the largest size of the general domain dataset, AIHub, the performance of ASR models trained from only AIHub is very poor. We conjecture this poor performance results from the differences between open domain and goal-oriented dialog datasets as shown in Fig.~\ref{fig:graph_stat}. On the other hand, when pretrained with AIHub and finetuned with QA Call or \prgname{}, both models show remarkable improvement. This supports the necessity of using task-specific data for ASR models in real-world goal-oriented services. 

In from-scratch training, both DS2 and LAS perform much better in the same domain than the different domain. Because, if a domain shifts, its data distribution and vocabulary also change. Moreover, QA Call provides more stable and better ASR performance than \prgname{} as well as in pretraning-finetuning scheme. We conjecture that these results are from larger size of QA Call testset. Also, QA Call contains a little more vocabulary and topics even though both speech corpora belong to goal-oriented dialog category.

In data augmentation experiments, no meaningful gain was found. We conjecture that two Call datasets have already been distorted by noises in the recording stage. In particular, the poor result of LAS on \prgname{} with SA is likely that too enhanced noise-based regularization harms the model capability of LAS with smaller parameter size, even if overall performance patterns of both models are similar to each other. 

These results confirm that task-specific speech corpora play a crucial role in improving ASR models for real-world goal-oriented dialog services such as \aicc{}. Therefore, we expect that our \prgname{} can considerably contribute to call-based reservation services. In addition, we can find that it is required to learn effective representation by pretraining with general-domain data to improve task-specific ASR models as well.

\section{Concluding Remarks}
We release a large-scale Korean goal-oriented dialog speech corpus, i.e., \prgname{}, which is useful for AI for Contact Center (\aicc) services. To the best of our knowledge, our dataset is the first Korean goal-oriented dialog speech corpus. Our \prgname{} contains 60,746 short sentence and utterance pairs under a restaurant reservation scenario. We verify the effectiveness of \prgname{} under three training schemes such as pretraining-finetuning, from-scratch learning, and data augmented from-scratch learning with two additional speech corpora. Experimental results support \prgname{} remarkably improves the performance of ASR models, thus being crucial for call-based restaurant reservation services. Furthermore, our \prgname{} can contribute to ASR models for diverse call-based reservation services, considering that many reservation services share common expressions such as working time and availability. 

The contribution of ClovaCall might be considered to be marginal because Korean is not a major language. On the contrary, however, opening low-resource language speech corpora to the public can enhance research diversity. Furthermore, \prgname{} is a goal-oriented dialog speech, which is rare in even major languages. Our dataset can contribute to design and construct other goal-oriented speech corpora in various languages.

\section{Acknowledgement}
The authors thank all members of Clova AI for supporting this work. In particular, we appreciate DUET, \aicc{}, and Speech teams including Icksang Han for data preparation and insightful discussion.

\bibliographystyle{IEEEtran}

\bibliography{references}


\end{document}